\documentclass[10pt,twocolumn,letterpaper]{article}

\usepackage[pagenumbers]{wacv} 

\definecolor{wacvblue}{rgb}{0.21,0.49,0.74}
\usepackage[pagebackref,breaklinks,colorlinks,allcolors=wacvblue]{hyperref}

\usepackage{multirow}
\usepackage{array} 
\usepackage[table,xcdraw]{xcolor}
\usepackage{algorithm}
\usepackage{algpseudocode}
\usepackage{kotex}

\def\wacvPaperID{938} 
\def\confName{WACV}
\def\confYear{2027}

\title{CILA: Confusion-based Instance-wise Logit Adjustment \\ for Long-Tailed Recognition}

\author{Jiyoung Lim\\
Sungkyunkwan University\\
{\tt\small maya0707@skku.edu}
\and
Jee-Hyong Lee \thanks{Corresponding author.} \\
Sungkyunkwan University\\
{\tt\small john@skku.edu}
}

\begin{document}

\maketitle
\begin{abstract}
Real-world datasets are inherently class-imbalanced, leading models to favor majority classes and struggle with minority ones. Existing logit adjustment methods address this by modifying logits based on class frequencies, but their class-wise adjustments ignore instance-wise confusion and often suppress informative signals from confusing non-target classes. This results in insufficient learning of such classes, increased prediction uncertainty, and limited generalization.

To address this, we propose two complementary methods.
First, \textbf{Confusion-based Instance-wise Logit Adjustment (CILA)} adaptively adjusts logits based on instance-wise confusion between the target and non-target classes, restoring informative gradients that class-wise adjustment suppresses. Second, \textbf{Batch-adaptation Noise-Injection (BNI)} perturbs the target logit during training, extending the confusion-aware update to correctly classified but near-boundary samples that the penalty alone cannot reach.
CILA is a loss-agnostic plug-in module that can be seamlessly integrated into existing long-tailed learning methods, and consistently improves performance across multiple benchmarks. 
Our code will be available at https://github.com/maya0395/CILA.
\end{abstract}
    
\section{Introduction}
\label{sec:intro}
Deep Neural Networks (DNNs) have achieved remarkable success, largely driven by the availability of well-curated and balanced datasets~\cite{krizhevsky2009learning, russakovsky2015imagenet}. 
However, real-world data often exhibit naturally imbalanced class distributions~\cite{everingham2010pascal, lin2014microsoft, van2017inaturalist}, 
where a few head classes contain abundant samples, while most tail classes have only limited data.
Training models on such long-tailed datasets poses significant challenges, as models tend to focus heavily on head classes while overlooking tail classes. 
Consequently, they become overly biased toward head classes, struggle to generalize to tail classes, and suffer from degraded overall performance.

To address the challenges of long-tailed distributions, various strategies have been proposed. 
Data re-balancing techniques~\cite{chawla2002smote, liu2008exploratory, shi2023re, cui2019class, tan2020equalization, kubat1997addressing, han2005borderline, he2009learning, barandela2003restricted, buda2018systematic, shen2016relay} mitigate class imbalance through resampling or reweighting. However, such methods often introduce instability and overfitting, especially under severe imbalance conditions\cite{cui2019class, huang2016learning, wang2023unified}. 
In contrast, logit adjustment techniques~\cite{menon2020long, ren2020balanced, son2024difficulty, cao2019learning, ye2020identifying, tan2020equalization, kini2021label} address the problem from a logit imbalance perspective by adding class-dependent adjustment terms to the logits.
This suppresses head class logits and amplifies tail classes, alleviating logit imbalance.

While effective, existing LA methods have a fundamental limitation:
they apply class-dependent adjustments uniformly to all logits. Such uniform adjustment reduces the logits of all classes, with tail classes experiencing a significantly larger reduction than head classes due to their lower frequency. 
For target logits, this relative difference helps mitigate logit imbalance by emphasizing the contribution of tail classes during training. However, when applied to non-target logits, the same adjustment uniformly suppresses their magnitudes according to class frequency, distorting the learning signals from confusing non-target classes. As a result, the model fails to capture meaningful distinctions among these confusing classes, making it difficult to learn precise and discriminative decision boundaries.

\begin{figure}[t] 
\begin{center}
\includegraphics[width=1\linewidth]{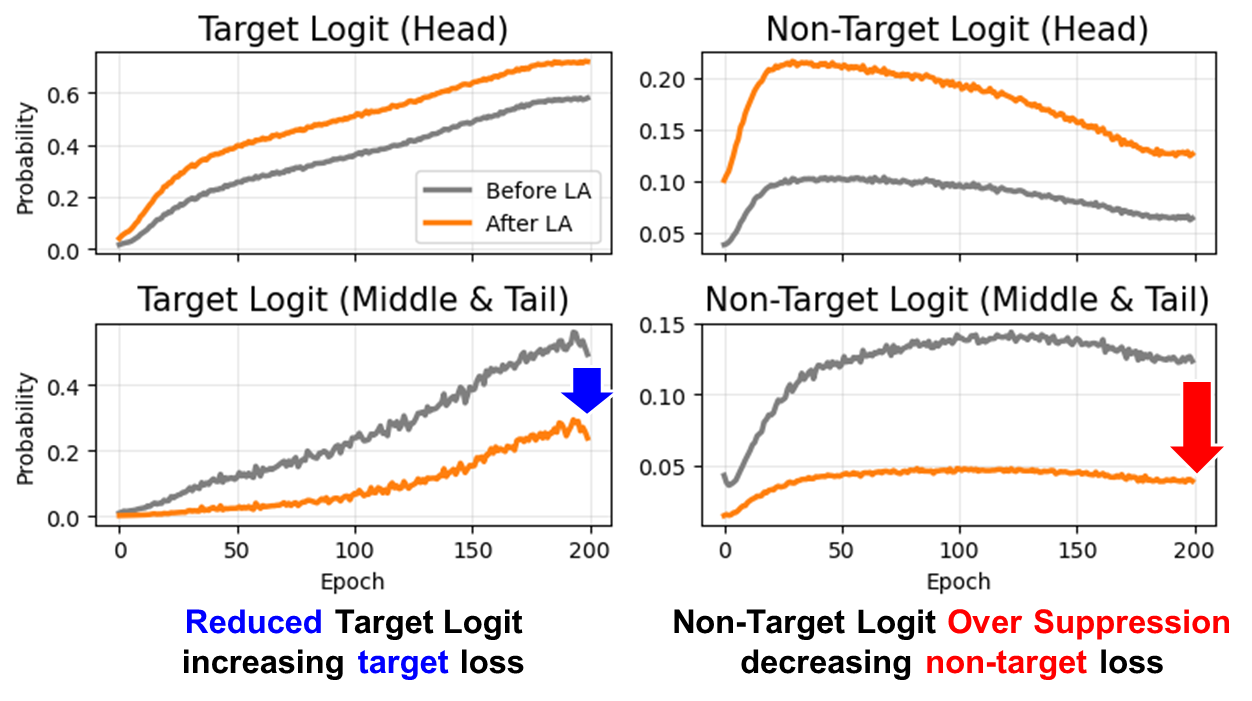}
\end{center}
\vspace{-1em}
\caption{Visualization of the Non-Target Logit Over-Suppression Problem on CIFAR100-IF100.
Logit adjustment increases head class target probabilities while reducing those of tail classes. However, it also \textbf{excessively suppresses non-target probabilities} of middle and tail classes, weakening informative gradients from confusing classes and hindering discrimination between confusing classes.}
\label{fig:vanishing}
\vspace{-1em}
\end{figure}

To further investigate this phenomenon, we analyze epoch-wise changes in the target and non-target probabilities of head, middle, and tail classes, as illustrated in Figure~\ref{fig:vanishing}. The gray and orange lines represent values before and after logit adjustment, respectively. As shown in the left column, the adjustment increases target probabilities of head classes while decreasing those of tail classes, appropriately shifting the learning emphasis toward tail samples. However, in the non-target case (right column), the same adjustment uniformly suppresses the non-target probabilities of middle and tail classes throughout training, weakening the gradient signal from confusing non-target categories.
We define this as the \textbf{Non-Target Logit Over-Suppression (NTLOS)} problem, where uniform logit adjustment excessively suppresses confusing non-target logits, reducing informative gradients and hindering fine-grained discrimination. Consequently, models trained under NTLOS exhibit higher prediction entropy on test data, as suppressed non-target representations fail to generalize to semantically similar classes.

To mitigate this issue, we propose two complementary methods.
\textbf{Confusion-based Instance-wise Logit Adjustment (CILA)}
dynamically adjusts each instance's logits through a confusion-based penalty, where confusion refers to cases in which one or more non-target logits exceed the target logit. This penalty preserves informative gradients from confusing non-target classes that uniform adjustment suppresses. 
However, the confusion penalty activates only for misclassified instances, leaving correctly classified but near-boundary samples unaddressed.
To complement this, we propose \textbf{Batch-adaptation Noise-Injection (BNI)}, which perturbs the target logit during training so that near-boundary correct samples become confusing after noise injection, making them eligible for the confusion-aware learning process. By simulating uncertainty in correct predictions, BNI encourages the model to maintain a wider decision margin, leading to more stable and flexible
decision boundaries.

CILA is loss-agnostic and can be seamlessly integrated into various long-tailed learning frameworks. 
Extensive experiments demonstrate that CILA consistently improves classification accuracy and training stability across multiple benchmarks, including CIFAR-100-LT, ImageNet-LT, and iNaturalist 2018.

The key contributions of this paper are summarized as follows:
\begin{itemize} 
\item We identify the Non-Target Logit Over-Suppression as a key limitation of existing logit adjustment methods.
\item We propose \textbf{CILA}, which applies a confusion-based
penalty to restore informative gradients from confusing non-target
classes.
\item We propose \textbf{BNI}, which extends the confusion-aware
update to near-boundary samples via batch-adaptive noise injection.
\end{itemize}

\begin{figure*}[ t!] 
\begin{center}
\includegraphics[width=0.8\linewidth]{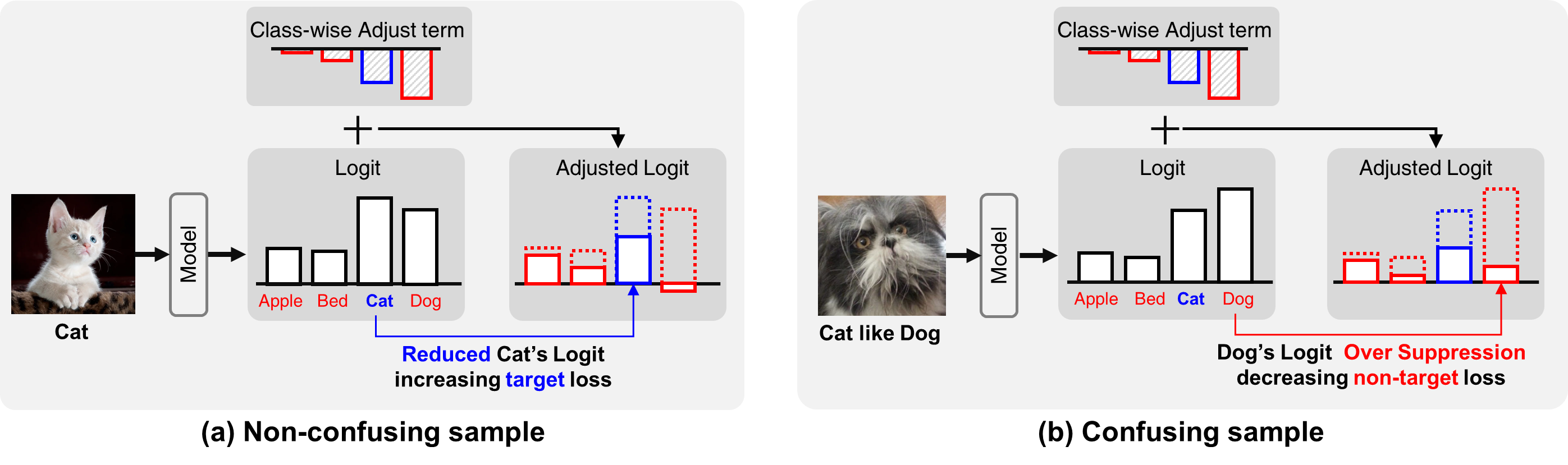}
\end{center}
\vspace{-1em}
\caption{Illustration of class-wise logit adjustment in non-confusing and confusing cases. (a) shows a non-confusing case that is easily distinguished. However, (b) represents a confusing case that is easily confused with another class (\texttt{Dog}). Class-wise logit adjustment increases a positive sample loss for the correct class while simultaneously reducing negative sample losses for confusing classes.}
\label{fig:problem}
\vspace{-1em}
\end{figure*}


\section{Related Work}
\textbf{Data Re-Balancing.}
Traditional approaches to address the class imbalance problem include Re-Sampling~\cite{ kubat1997addressing,chawla2002smote,han2005borderline,he2009learning,barandela2003restricted,buda2018systematic,shen2016relay,cao2019learning,shi2023re} or Re-Weighting~\cite{cui2019class,cao2019learning}. 
Re-sampling balances class distribution by increasing or decreasing the number of samples per class, while re-weighting assigns loss weights based on class frequencies to emphasize tail classes. Although effective, re-sampling can cause overfitting in tail classes or under-represent head classes, while re-weighting may cause unstable training due to excessively large gradients in tail classes.

\noindent\textbf{Logit Adjustment.}
To overcome data-level limitations, logit adjustment methods modify the classifier's output logits. 
Margin-based approaches~\cite{cao2019learning, zhao2022adaptive,son2024difficulty} enhance class separability by imposing instance-adaptive margins on target logits, yet still modify only target logits and thus fail to address the learning signal from confusing non-target classes. 
Logit Adjustment (LA)~\cite{ren2020balanced,menon2020long} applies class-frequency-derived terms to all logits, suppressing head-class dominance and emphasizing tail classes, and has been widely adopted across long-tailed recognition~\cite{zhu2022balanced,du2023global,li2022long}, zero-shot learning~\cite{chen2022zero}, and generalized classification~\cite{zhu2023generalized}.
However, both margin-based methods and LA share a fundamental limitation: they apply class-level adjustments uniformly across all instances, ignoring the instance-wise degree of confusion between the target and non-target classes. This causes NTLOS, suppressing informative gradients from confusing non-target classes regardless of how confusing each instance is.

\noindent\textbf{Confusion-aware Learning.}
CP-Mix~\cite{yoon2024mix} estimates class-pair confusion from the model's confusion matrix and synthesizes augmented samples between confusing class pairs, while OTMix~\cite{gao2023enhancing} uses optimal transport to generate semantically meaningful minority samples guided by class-level confusion. Both methods address confusion at the class level through data synthesis, relying on accumulated confusion statistics rather than the confusion of each individual instance. 

In contrast, CILA operates directly at the logit level, applying a penalty to each non-target logit based on its instance-wise degree of confusion with the target at every training step, without requiring data synthesis or a precomputed confusion matrix. 
Furthermore, BNI extends this update to correctly classified but near-boundary samples, which confusion-based augmentation methods do not address.

\section{Problem Setup}
\subsection{Logit Imbalance in Long-tailed Recognition}
\label{sec:problem_lt}
In long-tailed datasets, the class imbalance causes head-class logits to grow significantly larger during training, resulting in severe logit imbalance. To alleviate this, logit adjustment methods apply class-dependent terms derived from class frequencies to all logits:
\begin{equation}
\label{eqn:adjust_term}
\pi^{LA}_k = \log\Big(\frac{n_{k}}{\sum n_{j}}\Big),
\end{equation}
\begin{equation}
L_{LA} = -  \log \left( \frac{ e^{z_y+\pi^{LA}_y   }  }
{
\sum_{k=1}^N e^{z_k +\pi^{LA}_k  }   } \right)
\end{equation}
where $y$ denotes the ground truth (target) class for the given sample $x$. $z_k$ is the logit of class $k$, and $n_k$ is the number of training samples in class $k$.  

Figure~\ref{fig:problem} illustrates how logit adjustment operates in a four-class scenario (\texttt{Apple}, \texttt{Bed}, \texttt{Cat}, \texttt{Dog}), where \texttt{Cat} and \texttt{Dog} are tail classes. As shown in Figure~\ref{fig:problem}(a), the larger adjustment terms assigned to tail classes lower their logits (blue dashed $\to$ solid), helping the model focus more on tail classes and mitigating the logit imbalance between head and tail classes.

\subsection{Non-Target Logit Over-Suppression Problem}
\label{sec:problem_NSD}
While class-wise logit adjustment effectively mitigates class
imbalance for target logits, it carries a critical limitation
on non-target logits. Figure~\ref{fig:problem}(b) illustrates
this in a confusing case, where the input is a
\texttt{Cat-like-Dog} image easily mistaken for \texttt{Dog}.
The same adjustment is applied uniformly to all non-target
logits, reducing the logit for \texttt{Dog} (red dashed
$\to$ solid box) and lowering its loss contribution,
weakening the learning signal needed to differentiate
semantically similar categories. We term this the
\textbf{Non-Target Logit Over-Suppression (NTLOS)} problem,
where uniform logit adjustment excessively suppresses
confusing non-target logits, reducing informative gradients
and hindering fine-grained discrimination. We formally
characterize NTLOS at the gradient level in
Section~\ref{sec:grad_analysis}.

\section{Proposed Method}
As shown in Section~\ref{sec:problem_NSD}, uniform logit adjustment suppresses the gradient from confusing non-target classes, hindering fine-grained discrimination.
To address this, we propose two complementary methods.
First, \textbf{Confusion-based Instance-wise Logit Adjustment (CILA)} applies a confusion-based penalty proportional to the degree to which each non-target class challenges the target, preventing the over-suppression of informative non-target logits.
Second, \textbf{Batch-adaptation Noise-Injection (BNI)} extends the confusion-aware update to correctly classified but near-boundary samples, which the penalty alone cannot reach.

\subsection{Confusion-based Instance-wise LA (CILA)}
\label{sec:CILA}

We introduce Confusion-based Instance-wise Logit Adjustment, a method that adaptively adjusts each instance's logits based on the degree of confusion, quantified by how much a non-target logit $z_k$ exceeds the target logit $z_y$.
The adjustment is computed in three steps: (1) measuring a confusion-based penalty term, (2) weighting classes to reflect class imbalance, and (3) combining these to produce the final adjustment term.
 
\noindent\textbf{Step 1. Confusion-based Penalty Term.} 
We define class $k$ as confusing for a given instance when its logit exceeds the target logit, $z_k > z_y$.
To quantify the degree of confusion, we measure the relative logit magnitude anchored at $z_{\min}$:
\begin{equation}
\mathcal{C}_k = \frac{z_k - z_{\min}}{ z_y - z_{\min} + \epsilon},
\end{equation}
where $\epsilon$ is a small constant for numerical stability.
This formulation is motivated by the shift-invariance of the softmax, where logits can be expressed relative to $z_{min}$. 
Thus, $C_k$ provides a scale-normalized measure of how closely a non-target class approaches the target. In particular, $C_k > 1$ if and only if $z_k > z_y$, which is precisely our confusion criterion.

Based on $\mathcal{C}_k$, we define the confusion-based penalty as:
\begin{equation}
\label{eqn:penalty}
P_k = 
\operatorname{clip}~\!\Bigl(
\mathcal{C}_k - 1,\, 0,\; 2 \Bigr).
\end{equation}
Subtracting 1 activates the penalty only for genuinely confusing classes, while its magnitude increases with the degree of confusion to provide stronger adjustment to more strongly confused classes.
To prevent excessively large corrections from dominating optimization, we set the clipping bound to 2 based on the ablation in Table~\ref{tab:p_cliping}: smaller bounds discard informative signals for tail samples, whereas a bound of 2 yields the largest tail-class gains.




\noindent \textbf{Step 2. Class Weighting Term.}
In long-tailed distributions, tail classes rarely encounter confusing samples, though such instances are crucial for refining decision boundaries.
To address this, we modulate the confusion penalty using class-wise weights, which amplify the penalty when the target class is a tail class. 
Following prior studies~\cite{park2022cmo}, the class-wise weight is defined as follows: 
\begin{equation}
\label{eqn:class_wise}
D_k \;=\; \frac{1/n_k}{\sum_j 1/n_j}\, \cdot  C  
\end{equation}
where $n_k$ denotes the number of training instances in class $k$, and $C$ is the total number of classes. 
The normalized inverse-frequency weight ensures stronger penalties for tail classes ($n_k$ small).
 
\noindent \textbf{Step 3. Confusion-based Logit Adjustment.}
The final logit adjustment term for each class $k$ is computed by combining the confusion penalty and the Class Weighting term:
\begin{equation}
\label{eqn:instance_wise}
\pi_k^{C} = P_k \cdot D_y,
\end{equation}
where $D_y$ is the weight of the target class $y$. 
This formulation amplifies the penalty for confusing non-targets when the target is a tail class, encouraging the model to resolve instance-level ambiguity.

The confusion-aware logit adjustment is incorporated into the training loss as follows:
\begin{equation}
\label{eqn:CILA}
L_{CILA} = -  \log \left( \frac{ e^{z_y+\pi^{C}_y }  }
{ \sum_{k=1}^N e^{z_k +\pi^{C}_k }   } \right).
\end{equation}

\begin{table*}[t!]
\scriptsize
\centering
\caption{Performance comparison with loss function-based on the CIFAR-100-LT, re-implemented from official code (CUDA/DODA/SHIKE: respective repositories; others: DBM~\cite{son2024difficulty})}
\renewcommand{\arraystretch}{1.3}
\vspace{-1em}
\resizebox{1\textwidth}{!}{%
\begin{tabular}{ll | cccc | cccc | cccc}
\toprule
& &
\multicolumn{4}{c|}{IF 10} &
\multicolumn{4}{c|}{IF 50} &
\multicolumn{4}{c}{IF 100} \\ \cline{3-14}
Method & Loss &
Head & Med & Tail & All &
Head & Med & Tail & All &
Head & Med & Tail & All \\ \hline

\multirow{11}{*}{Loss} &
CE &
\textbf{67.4$\pm$0.5} & 45.7$\pm$0.5 & - & 60.7$\pm$0.5 &
69.7$\pm$0.4 & 40.2$\pm$0.6 & 18.0$\pm$0.7 & 48.3$\pm$0.3 &
\textbf{72.0$\pm$0.3} & 43.5$\pm$0.6 & 10.0$\pm$0.4 & 43.4$\pm$0.3 \\
& LDAM &
67.1$\pm$0.2 & 46.5$\pm$0.8 & - & 60.8$\pm$0.4 &
\textbf{69.8$\pm$0.3} & 39.3$\pm$0.5 & 16.3$\pm$0.5 & 47.7$\pm$0.0 &
71.4$\pm$0.4 & 42.4$\pm$1.1 & 10.5$\pm$0.5 & 43.0$\pm$0.4 \\
& DBM &
67.7$\pm$0.5 & 46.8$\pm$0.7 & - & 61.2$\pm$0.4 &
69.8$\pm$0.2 & 40.0$\pm$0.5 & 17.7$\pm$0.5 & 48.2$\pm$0.2 &
71.7$\pm$0.8 & 42.3$\pm$0.5 & 10.2$\pm$0.4 & 43.0$\pm$0.2 \\
\rowcolor[HTML]{D9E2F3} \cellcolor{white} & CILA &
66.8$\pm$0.1 & \textbf{54.1$\pm$0.3} & - & \textbf{62.7$\pm$0.3} &
68.0$\pm$0.2 & \textbf{44.6$\pm$0.5} & \textbf{29.0$\pm$0.4} & \textbf{51.2$\pm$0.3} &
70.3$\pm$0.5 & \textbf{46.1$\pm$0.2} & \textbf{19.2$\pm$0.8} & \textbf{46.1$\pm$0.1} \\
\cline{2-14}

& CE-DRW &
65.1$\pm$0.7 & 54.2$\pm$1.6 & - & 61.7$\pm$0.2 &
65.9$\pm$0.7 & 44.9$\pm$1.0 & 28.4$\pm$0.5 & 50.5$\pm$0.1 &
67.6$\pm$0.9 & 47.3$\pm$0.8 & 19.0$\pm$0.3 & 45.9$\pm$0.2 \\
& LDAM-DRW &
61.7$\pm$1.9 & 59.9$\pm$3.0 & - & 61.2$\pm$0.3 &
56.8$\pm$0.9 & 49.2$\pm$0.1 & 39.4$\pm$0.3 & 50.5$\pm$0.3 &
55.9$\pm$0.7 & \textbf{49.6$\pm$0.2} & 31.8$\pm$0.4 & 46.5$\pm$0.1 \\
& DBM-DRW &
63.6$\pm$0.1 & \textbf{55.2$\pm$0.4} & - & 61.0$\pm$0.1 &
66.1$\pm$0.2 & 46.3$\pm$0.2 & 29.0$\pm$0.6 & 51.3$\pm$0.1 &
67.3$\pm$0.6 & 47.8$\pm$0.4 & 19.9$\pm$0.7 & 46.2$\pm$0.2 \\
\rowcolor[HTML]{D9E2F3} \cellcolor{white} & CILA-DRW &
\textbf{66.5$\pm$0.1} & 55.0$\pm$0.3 & - & \textbf{62.7$\pm$0.2} &
\textbf{67.0$\pm$0.6} & \textbf{46.2$\pm$0.1} & \textbf{29.6$\pm$0.6} & \textbf{51.5$\pm$0.2} &
\textbf{68.1$\pm$1.1} & 48.0$\pm$0.5 & \textbf{20.4$\pm$0.4} & \textbf{46.5$\pm$0.2} \\
\cline{2-14}

& LA &
65.6$\pm$0.3 & 55.2$\pm$0.4 & - & 62.4$\pm$0.3 &
\textbf{65.2$\pm$0.3} & 46.9$\pm$0.3 & 37.5$\pm$1.4 & 52.5$\pm$0.2 &
65.1$\pm$0.3 & 47.9$\pm$0.7 & 27.5$\pm$0.5 & 47.8$\pm$0.3 \\
& LA + DBM &
65.4$\pm$0.4 & 54.8$\pm$0.4 & - & 62.1$\pm$0.2 &
\textbf{65.2$\pm$0.3} & 47.0$\pm$0.4 & 34.8$\pm$1.3 & 52.3$\pm$0.4 &
\textbf{66.6$\pm$0.1} & 48.1$\pm$0.6 & 25.4$\pm$0.73 & 47.8$\pm$0.3 \\
\rowcolor[HTML]{D9E2F3} \cellcolor{white} & LA+CILA &
\textbf{65.9$\pm$0.1} & \textbf{57.1$\pm$0.1} & - & \textbf{63.2$\pm$0.1} &
63.5$\pm$0.1 & \textbf{48.5$\pm$0.5} & \textbf{41.3$\pm$0.9} & \textbf{52.6$\pm$0.3} &
64.4$\pm$0.2 & \textbf{49.1$\pm$0.8} & \textbf{30.0$\pm$0.6} & \textbf{48.5$\pm$0.5} \\
\hline

& LA &
64.8$\pm$0.1 & 58.1$\pm$0.8 & - & 62.7$\pm$0.2 &
60.7$\pm$0.6 & 48.2$\pm$1.3 & 41.0$\pm$0.8 & 52.0$\pm$0.7 &
59.5$\pm$0.8 & 49.6$\pm$0.3 & 32.2$\pm$0.2 & 47.8$\pm$0.1 \\
\rowcolor[HTML]{D9E2F3} \cellcolor{white}\multirow{-2}{*}{CMO} & LA+CILA &
\textbf{66.5$\pm$0.2} & \textbf{61.3$\pm$0.1} & - & \textbf{64.9$\pm$0.2} &
\textbf{62.3$\pm$1.2} & \textbf{50.7$\pm$1.3} & \textbf{46.5$\pm$0.4} & \textbf{54.7$\pm$0.1} &
\textbf{61.0$\pm$0.4} & \textbf{50.1$\pm$0.8} & \textbf{36.4$\pm$1.1} & \textbf{49.8$\pm$0.2} \\
\hline

& LA &
66.3$\pm$0.1 & 55.9$\pm$0.1 & - & 63.1$\pm$0.1 &
66.2$\pm$1.3 & 47.5$\pm$0.5 & 35.9$\pm$0.3 & 53.1$\pm$0.7 &
66.8$\pm$0.5 & 48.8$\pm$0.9 & 27.2$\pm$0.3 & 48.6$\pm$0.5 \\
\rowcolor[HTML]{D9E2F3} \cellcolor{white}\multirow{-2}{*}{BCL} & LA+CILA &
\textbf{68.2$\pm$0.3} & \textbf{59.2$\pm$0.1} & - & \textbf{65.4$\pm$0.3} &
\textbf{67.0$\pm$0.3} & \textbf{52.9$\pm$0.2} & \textbf{43.9$\pm$0.2} & \textbf{57.1$\pm$0.0} &
\textbf{68.6$\pm$1.4} & \textbf{53.2$\pm$0.1} & \textbf{33.7$\pm$0.3} & \textbf{52.8$\pm$0.5} \\
\hline

& LA &
68.0$\pm$0.2 & \textbf{59.2$\pm$0.9} & - & \textbf{65.6$\pm$0.2} &
68.8$\pm$0.2 & 52.1$\pm$0.3 & 42.1$\pm$0.9 & 57.3$\pm$0.3 &
69.4$\pm$0.2 & 54.6$\pm$0.3 & \textbf{31.8$\pm$0.1} & 53.1$\pm$0.0 \\
\rowcolor[HTML]{D9E2F3} \cellcolor{white}\multirow{-2}{*}{GML} & LA+CILA &
\textbf{69.4$\pm$0.2} & 56.8$\pm$1.5 & - & 65.5$\pm$0.4 &
\textbf{69.1$\pm$0.2} & \textbf{56.0$\pm$0.3} & 42.1$\pm$1.9 & \textbf{59.0$\pm$0.1} &
\textbf{70.4$\pm$0.5} & \textbf{57.3$\pm$0.6} & 31.4$\pm$0.1 & \textbf{54.2$\pm$0.4} \\
\hline

& LA &
67.6$\pm$0.3 & \textbf{60.2$\pm$0.5} & - & \textbf{65.3$\pm$0.2} &
68.4$\pm$0.4 & 53.8$\pm$0.0 & \textbf{46.9$\pm$1.0} & 58.6$\pm$0.3 &
69.3$\pm$0.4 & 55.9$\pm$0.3 & 34.8$\pm$0.6 & 54.3$\pm$0.2 \\
\rowcolor[HTML]{D9E2F3} \cellcolor{white}\multirow{-2}{*}{NCL} & LA+CILA &
\textbf{68.2$\pm$0.4} & 58.5$\pm$1.2 & - & 65.2$\pm$0.2 &
\textbf{68.5$\pm$0.1} & \textbf{55.2$\pm$0.4} & 45.0$\pm$1.8 & \textbf{58.8$\pm$0.2} &
\textbf{69.4$\pm$1.9} & \textbf{57.7$\pm$1.4} & \textbf{34.9$\pm$3.1} & \textbf{55.0$\pm$0.2} \\
\hline

& LA &
64.1$\pm$0.2 & 56.6$\pm$0.4 & - & 61.7$\pm$0.2 &
62.5$\pm$0.8 & 47.7$\pm$0.8 & 37.6$\pm$0.4 & 51.9$\pm$0.4 &
62.6$\pm$0.1 & 48.3$\pm$0.2 & 28.6$\pm$0.7 & 47.4$\pm$0.3 \\
\rowcolor[HTML]{D9E2F3} \cellcolor{white}\multirow{-2}{*}{CUDA} & LA+CILA &
\textbf{67.4$\pm$0.3} & \textbf{59.0$\pm$0.2} & - & \textbf{64.8$\pm$0.2} &
\textbf{64.8$\pm$0.6} & \textbf{51.3$\pm$0.5} & \textbf{42.0$\pm$0.1} & \textbf{55.2$\pm$0.2} &
\textbf{64.9$\pm$0.6} & \textbf{51.0$\pm$0.6} & \textbf{33.0$\pm$0.9} & \textbf{50.5$\pm$0.3} \\
\hline

& LA &
62.2$\pm$0.2 & 55.3$\pm$0.3 & - & 60.1$\pm$0.2 &
61.3$\pm$0.6 & 47.9$\pm$0.5 & 35.2$\pm$1.3 & 51.1$\pm$0.4 &
62.0$\pm$1.3 & 48.9$\pm$0.6 & 25.7$\pm$0.2 & 46.5$\pm$0.5 \\
\rowcolor[HTML]{D9E2F3} \cellcolor{white}\multirow{-2}{*}{DODA} & LA+CILA &
\textbf{65.5$\pm$0.3} & \textbf{58.1$\pm$1.5} & - & \textbf{63.2$\pm$0.3} &
\textbf{63.2$\pm$0.2} & \textbf{52.2$\pm$0.5} & 41.0$\pm$1.0 & \textbf{54.7$\pm$0.3} &
\textbf{64.2$\pm$1.3} & \textbf{51.5$\pm$1.2} & \textbf{32.0$\pm$0.8} & \textbf{50.1$\pm$0.2} \\
\hline

& LA &
70.0$\pm$0.2 & 61.9$\pm$0.4 & - & 67.5$\pm$0.3 &
\textbf{70.8$\pm$0.1} & 54.2$\pm$0.3 & 42.9$\pm$0.4 & 59.0$\pm$0.2 &
\textbf{71.5$\pm$0.3} & 56.7$\pm$0.2 & 35.8$\pm$0.2 & 55.6$\pm$0.1 \\
\rowcolor[HTML]{D9E2F3} \cellcolor{white}\multirow{-2}{*}{SHIKE} & LA+CILA &
\textbf{70.5$\pm$0.8} & \textbf{62.6$\pm$0.7} & - & \textbf{68.2$\pm$0.2} &
70.0$\pm$0.1 & \textbf{55.9$\pm$0.3} & \textbf{45.9$\pm$0.3} & \textbf{60.6$\pm$0.1} &
67.6$\pm$0.8 & \textbf{57.6$\pm$0.5} & \textbf{41.4$\pm$0.3} & \textbf{57.0$\pm$0.5} \\
\bottomrule
\end{tabular}%
}
\label{tab:results_all}
\vspace{-1em}
\end{table*}

\begin{figure}
\begin{center}
\includegraphics[width=1 \linewidth]{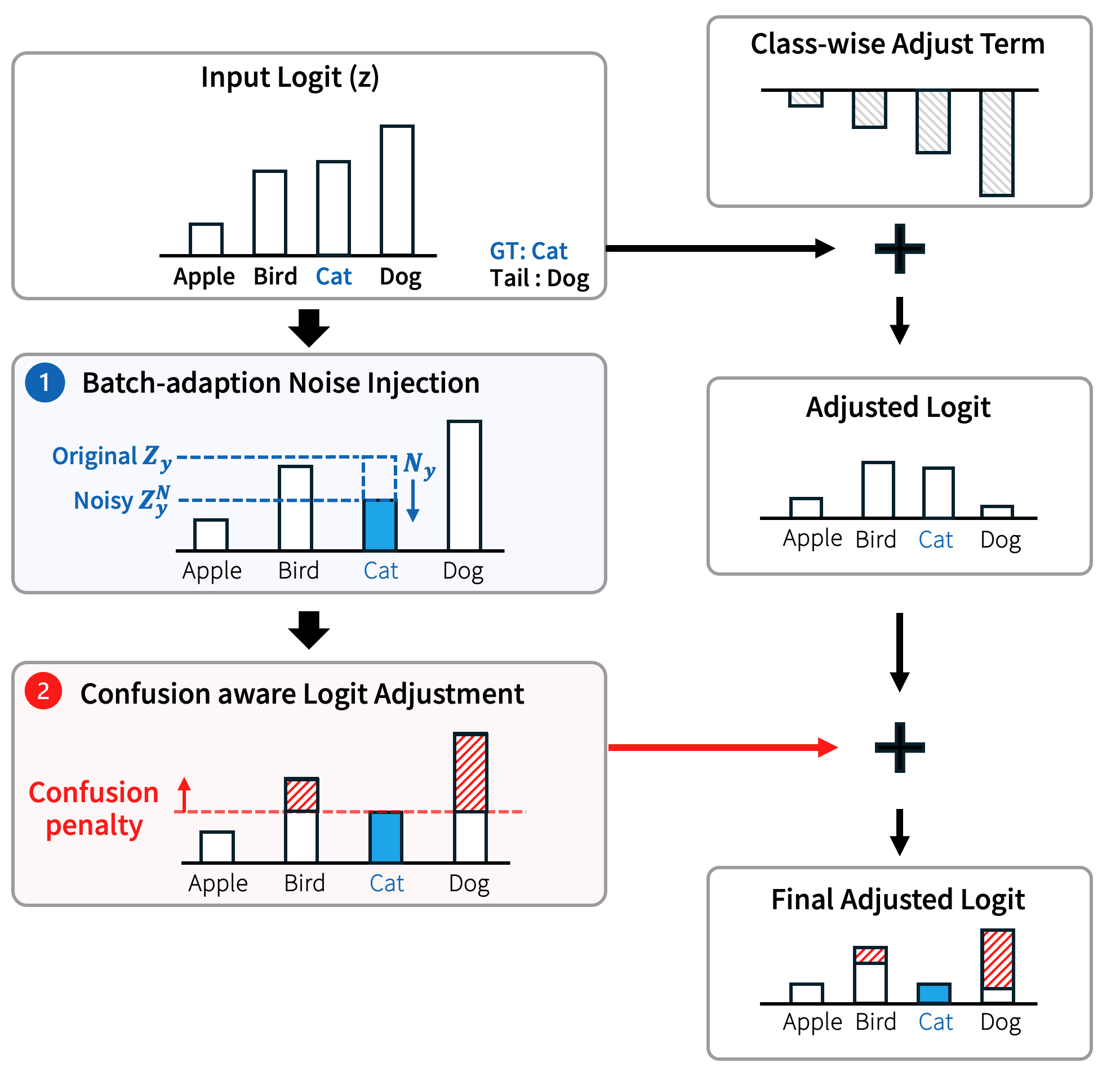}
\end{center}
\vspace{-2em}
\caption{Overview of the proposed method. 
}
\label{fig:method}
\vspace{-1em}
\end{figure}

\subsection{Batch-adaptation Noise-Injection (BNI)}
The confusion penalty activates only for misclassified instances ($z_k > z_y$), leaving correctly classified but near-boundary samples unaddressed. 
To complement this, we introduce \textbf{Batch-adaptation Noise-Injection (BNI)}, which perturbs the target logit so that $z_y^N$ replaces $z_y$ in Eq.~\ref{eqn:CILA} before computing $\mathcal{C}_k$, enabling CILA to handle near-boundary correct samples.

\noindent\textbf{Noise Injection.} To expose near-boundary correct samples to the confusion-aware penalty, BNI reduces the target logit by injecting a dynamically scaled noise term before computing $C_k$, so that the confusion penalty is evaluated on the perturbed target. The target logit $z_y$ is perturbed as:
\begin{equation}
  z^N_y = z_y - N_y, \qquad N_y = D_y \cdot \xi,
  \quad \xi \sim \mathrm{Rayleigh}(\sigma),
  \label{eq:nir}
\end{equation}
where $D_y$ is the class-dependent weight defined in Eq.~\ref{eqn:class_wise}, with larger $D_y$ for tail classes concentrating the regularization where decision boundaries are most uncertain. 

We adopt the Rayleigh distribution for $\xi$ because, unlike Gaussian whose density peaks at zero, Rayleigh attains its mode at $\sigma > 0$, ensuring non-trivial perturbation magnitude. Formally, its expected value satisfies:
\begin{equation}
  \mathbb{E}[\xi] = \sigma \cdot \sqrt{\frac{\pi}{2}} > 0,
  \label{eq:rayleigh}
\end{equation}
guaranteeing that the noise consistently induces a meaningful reduction in the target logit rather than being dominated by near-zero perturbations. The scale parameter $\sigma$ is not fixed but adapts dynamically to the current training state, as described next.

\noindent\textbf{Batch-Adaptive Noise Scale.}
Since logit magnitudes grow substantially during training, a fixed $\sigma$ is either too small in late epochs or too large in early ones. We therefore set $\sigma$ proportional to the batch mean target logit
$\bar{z}_y = \frac{1}{|B|}\sum_{(x,y)\in B} z_y$:
\begin{equation}
  \sigma = \bar{z}_y \cdot \rho,
  \label{eq:sigma}
\end{equation}
where $\rho < 1$ is a scaling factor derived from the stability condition detailed below. Since $\bar{z}_y$ can be near zero or negative during early training, we apply a minimum value of 0.5 in practice to ensure $\sigma > 0$ as required by the Rayleigh distribution.

Using $\bar{z}_y$ directly as $\sigma$ would collapse the target logit even for well-classified samples. We therefore impose the stability condition $\mathbb{E}[N_y] < \bar{z}_y$, which gives: 
\begin{equation}
  \mathbb{E}[N_y] = D_y \cdot \bar{z}_y \cdot \rho \cdot
  \sqrt{\frac{\pi}{2}} < \bar{z}_y,
\end{equation}
and yields the upper bound:
\begin{equation}
  \rho < \sqrt{\frac{2}{\pi}}\,\frac{1}{D_{\mathrm{tail}}},
  \label{eq:rho_condition}
\end{equation}
where $D_{\mathrm{tail}}$ is the weight of the most underrepresented class. For CIFAR-100-LT (IF=100), $D_{\mathrm{tail}} \approx 3.97$ gives $\rho \lesssim 0.201$, so we set $\rho = 0.2$. For other datasets, $\rho$ is determined analogously from Eq.~\ref{eq:rho_condition} using their respective $D_{\mathrm{tail}}$ values. 

Unlike existing noise regularization methods that primarily suppress overconfidence, BNI serves a fundamentally different purpose: it intentionally simulates uncertainty in correctly predicted but near-boundary samples, making them eligible for the confusion-aware penalty. Rather than smoothing confidence globally, BNI concentrates its effect on tail classes via the class-dependent scale $D_y$, adaptively expanding the set of informative instances that refine decision boundaries.

\begin{table*}[t]
\centering
\small
\renewcommand{\tabcolsep}{3mm}
\caption{ Accuracy (\%) on ImageNet-LT and iNaturalist 2018. † and ‡ denote results from DODA~\cite{wang2024doda} and DBM~\cite{son2024difficulty}, respectively.}
\begin{tabular}{l|cccc|cccc}
\toprule
 & \multicolumn{4}{c|}{ImageNet-LT} 
 & \multicolumn{4}{c}{iNaturalist 2018} \\ \cline{2-9}

Method
& Head & Med & Tail & All &
  Head & Med & Tail & All  \\ 
\hline
CE \textdagger
& 64.0 & 33.8 & 5.8 & 41.6
& \textbf{73.9} & 63.5 & 55.5 & 61.0 \\
cRT \textdagger
& 58.8 & 44.0 & 26.1 & 47.3
& 69.0 & 66.0 & 63.2 & 65.2 \\
LWS \textdagger
& 57.1 & 45.2 & 29.3 & 47.7
& 65.0 & 66.3 & 65.5 & 65.9 \\
CE-DRW \textdagger
& 61.7 & 47.3 & 28.8 & 50.1
& 68.2 & 67.3 & 66.4 & 67.0 \\
LDAM-DRW \textdagger
& 60.4 & 46.9 & 30.7 & 49.8
& - & - & - & 66.1 \\
LA \textdagger
& 60.9 & 48.8 & 32.1 & 51.0
& 65.5 & 67.5 & 67.5 & 67.2 \\
ALA
& 62.4 & 49.1 & 35.7 & 52.4
& 71.3 & 70.8 & 70.4 & 70.7 \\
LA + OTmix
& 64.0 & 52.4 & 42.7 & 55.6
& 71.1 & \textbf{72.0} & 70.8 & 71.5 \\
\rowcolor[HTML]{D9E2F3}
LA + CILA
 & 62.9$\pm$0.1 & 53.7$\pm$0.3 & 40.7$\pm$0.4 & 55.5$\pm$0.1
& 66.1$\pm$0.2 & 71.3$\pm$0.1 & 73.8$\pm$0.3 & 71.7$\pm$0.2 \\

\hline
BCL \textdagger
& 65.3 & 53.5 & 36.3 & 55.6
& 68.2 & 71.3 & 71.3 & 71.0 \\
BCL-DBM \textdaggerdbl
& - & - & - & 56.3
& 65.6 & 71.8 & 73.8 & 71.9 \\
\rowcolor[HTML]{D9E2F3}
BCL + CILA
& \textbf{65.6$\pm$0.1} & \textbf{55.4$\pm$0.3} & \textbf{44.3$\pm$0.4} & \textbf{57.8$\pm$0.1}
& 67.1$\pm$1.0 & 71.8$\pm$0.4 & \textbf{74.6$\pm$0.5} & \textbf{72.4$\pm$0.1} \\

\bottomrule
\end{tabular}
\vspace{-1em}
\label{tab:imagenet_inat_combined}
\end{table*}

\subsection{Gradient Analysis}
\label{sec:grad_analysis}
We formally compare the gradient flow under LA and CILA to
characterize NTLOS and show that CILA provably restores the
suppressed gradient.

\noindent\textbf{Gradient under LA.}
The gradient of $\mathcal{L}_{\mathrm{LA}}$ with respect to
a non-target logit $z_k$ ($k \neq y$) is $\tilde{p}_k$, where
$\tilde{p}_k = \frac{\exp(z_k + \pi^{\mathrm{LA}}_k)}
{\sum_j \exp(z_j + \pi^{\mathrm{LA}}_j)}$.
For a confusing tail class $k$, $\pi^{\mathrm{LA}}_k \ll 0$ substantially reduces $\tilde{p}_k$ regardless of instance-level confusion, confirming NTLOS as a gradient-level phenomenon:
\begin{equation}
  \frac{\partial \mathcal{L}_{\mathrm{LA}}}{\partial z_k}
  = \tilde{p}_k \;\ll\; p_k, \qquad
  \text{for confusing tail class } k,
  \label{eq:ntlos_grad}
\end{equation}
where $p_k = \frac{\exp(z_k)}{\sum_j \exp(z_j)}$ is the
standard softmax probability without adjustment. Thus, the
gradient signal from confusing non-target classes is
significantly suppressed by LA, regardless of how confusing
the instance actually is.

\noindent\textbf{Gradient Restoration by CILA.}
Under CILA, the gradient takes the same form with
$\tilde{q}_k = \frac{\exp(z_k + \pi^{C}_k)}
{\sum_j \exp(z_j + \pi^{C}_j)}$.
For a confusing class $k$ satisfying $z_k > z_y$,
the confusion penalty satisfies $\pi^{C}_k > 0 > \pi^{\mathrm{LA}}_k$,
so the adjusted logit of class $k$ is strictly larger under
CILA than under LA. For non-confusing classes $j$ where
$z_j \leq z_y$, $P_j = 0$ and thus $\pi^C_j = 0$, so
their adjusted logits remain unchanged. 
Since the softmax probability is monotonically increasing in its own logit, it follows that $\tilde{q}_k > \tilde{p}_k$, and therefore:
\begin{equation}
  \frac{\partial \mathcal{L}_{\mathrm{CILA}}}{\partial z_k}
  = \tilde{q}_k \;>\; \tilde{p}_k
  = \frac{\partial \mathcal{L}_{\mathrm{LA}}}{\partial z_k},
  \quad \forall\, k \text{ s.t. } z_k > z_y.
  \label{eq:grad_comparison}
\end{equation}
As $\pi^{C}_k$ increases with $z_k - z_y$, the gradient correction grows with the degree of confusion, meaning the most confusing instances receive the strongest learning signal.

Furthermore, BNI perturbs the target logit before computing the confusion penalty:
\begin{equation}
z_y^{N}=z_y-N_y,\qquad
N_y=D_y\xi,\quad \xi\sim\mathrm{Rayleigh}(\sigma).
\end{equation}
Since the noise scale is adapted to the current batch logit magnitude, the resulting perturbation follows the training state and exposes correctly classified but near-boundary samples to the same confusion-aware gradient mechanism.

\subsection{Stability of the Combined Objective.}
We analyze the stability of the combined CILA-BNI objective by separately controlling the magnitudes of the confusion-based correction and the injected perturbation. 

First, the CILA penalty is explicitly bounded by the clipping operation in Eq.~(4), such that $0 \leq P_k \leq 2$. Therefore, the additional logit correction satisfies $0 \leq \pi_k^C \leq 2D_y$, where $D_y$ is the class-dependent weight in Eq.~(5). This bound prevents the confusion-aware correction from growing arbitrarily large, even for highly confusing instances.
Second, BNI controls the magnitude of the target-logit perturbation through the batch-adaptive scale in Eq.~(10). Specifically, imposing $\mathbb{E}[N_y] < \bar{z}_y$ yields the upper bound on $\rho$ in Eq.~(12). Thus, the expected perturbation remains smaller than the batch-mean target logit, preventing the injected noise from dominating the target signal.

Together, these constraints bound the CILA and BNI corrections, enabling confusion-aware gradient correction while preventing unbounded logit modifications and supporting stable optimization.


\section{Experiments}
\label{sec:experiment}
\noindent
\textbf{Datasets.} 
To ensure fair comparisons, we conduct experiments on three widely used long-tailed image classification benchmarks: CIFAR-100-LT~\cite{cao2019learning}, ImageNet-LT~\cite{liu2008exploratory}, and iNaturalist 2018~\cite{van2017inaturalist}.

\noindent
\textbf{Comparison Baselines.} We evaluate our proposed method against a diverse set of baseline methods to ensure fair and comprehensive comparisons across different learning strategies for long-tailed recognition. The baselines include:  
\begin{itemize}
    \item Loss Function-Based Method: Cross-Entropy~(CE)~\cite{he2016resnet}
    \item Re-weighting Method: DRW~(NeurIPS'19)~\cite{cao2019learning}
    \item Logit Adjustment Method: LDAM~(NeurIPS'19)~\cite{cao2019learning}, LA~(ICLR'21)~\cite{menon2020long}, ALA~(AAAI'22)~\cite{zhao2022adaptive}, and DBM (AAAI'25)~\cite{son2024difficulty}
    \item Augmentation Method: CMO~(CVPR'22)~\cite{park2022cmo}, OTmix~(NeurIPS'23)~\cite{gao2023enhancing}
    \item Curriculum Learning Method: CUDA~(ICLR'23)~\cite{ahn2023cuda}, DODA~(ICLR'24)~\cite{wang2024doda}
    \item Contrastive Learning Method: BCL~(CVPR'22)~\cite{zhu2022balanced}   
    \item Mutual Information Maximization Method: GML~(ICML'23)~\cite{gmlsuh2023}
    \item Ensemble-based method: NCL~(CVPR'22)~\cite{nclli2022nested}, SHIKE~(CVPR'23)~\cite{shikejin2023long} 
\end{itemize}
In the tables, \textbf{CILA} denotes our full method comprising Confusion-based Logit Adjustment and Batch-adaptation Noise-Injection (BNI). As a plug-in module, CILA can be integrated into existing methods and is evaluated both as a standalone method and in combination with representative baselines.


\noindent \textbf{Implementation Details.} 
Following the experimental setup of prior work~\cite{cao2019learning, menon2020long, son2024difficulty}, all implementation details are provided in the supp.~3.

\subsection{Experimental Results}
\noindent\textbf{CIFAR-100-LT.} 
 Table~\ref{tab:results_all} presents the performance comparison on CIFAR-100-LT under varying imbalance factors (IF=10, 50, 100). CILA consistently improves over loss function-based, re-weighting, and logit adjustment baselines, and further generalizes as a plug-in module across a range of representative long-tailed learning methods spanning diverse learning paradigms.


\begin{figure*}[t] 
\begin{center}
\includegraphics[width=1\linewidth]{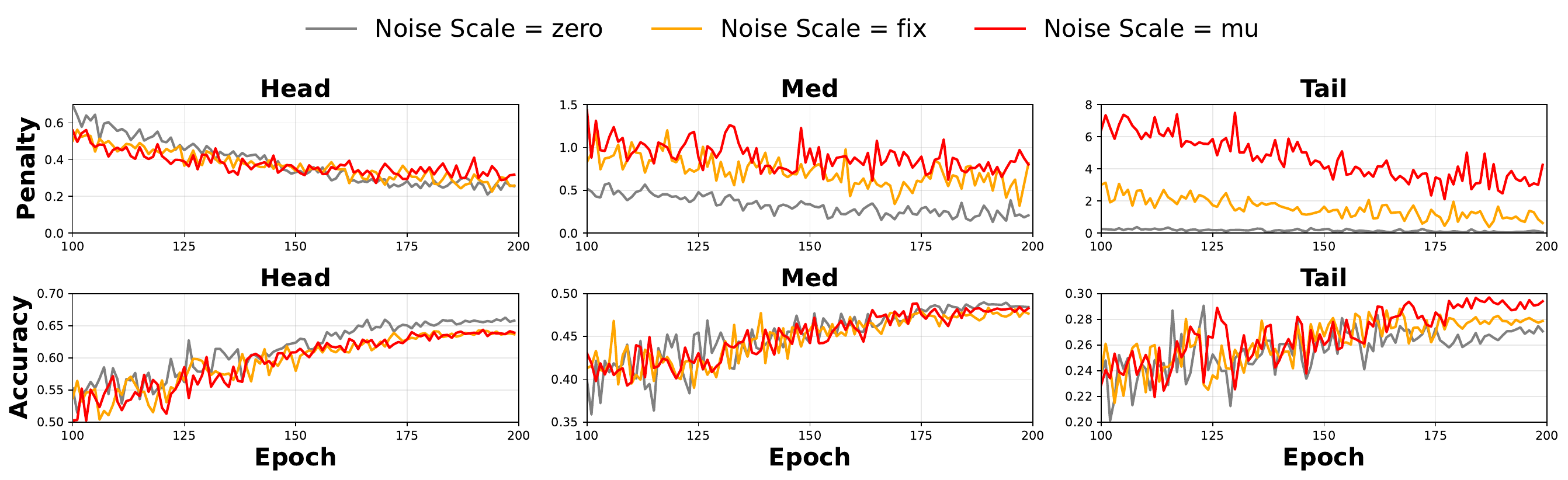}
\end{center}
\vspace{-1em}
\caption{Confusion penalty and accuracy across 
training epochs under three noise settings: no noise~(zero), fixed-scale noise~(fix), and batch-adaptive noise~(mu) (CIFAR-100-LT, IF=100).}
\vspace{-1em}
\end{figure*}

\noindent \textbf{ImageNet-LT and iNaturalist 2018.} 
Table~\ref{tab:imagenet_inat_combined} presents the performance of the proposed method on ImageNet-LT and iNaturalist 2018 datasets. 
BCL+CILA achieves the best performance on both datasets, with particularly notable gains in tail classes, demonstrating that the confusion-aware penalty and batch-adaptive noise injection effectively improve discrimination in rare categories.

\subsection{Analysis of Proposed Method}
We analyze the effectiveness of CILA and BNI through four complementary
evaluations: feature discriminability, prediction uncertainty,
confusion between semantically similar classes, and the effect of
batch-adaptive noise scaling.

\noindent\textbf{Feature Discriminability.}
We measure intra-class compactness and inter-class separability following~\cite{son2024difficulty}. As shown in Table~\ref{tab:dbm_analysis}, CILA-LA achieves smaller intra-class distances and higher inter-class separability than LA and DBM-LA across all class groups. Unlike DBM-LA, which improves separability at the cost of increased intra-class distances, CILA-LA improves both metrics simultaneously, demonstrating more discriminative representations.

\begin{table}[h]
\centering
\scriptsize
\setlength{\tabcolsep}{4.5pt}
\caption{Analysis of inter-class separability and intra-class compactness on CIFAR100~(IF 100).}
\vspace{-1em}
\begin{tabular}{l | cccc| cccc  }
\toprule
\multirow{2}{*}{Method} 
& \multicolumn{4}{c |}{Inter-class Separability $\uparrow$} 
& \multicolumn{4}{c  }{Intra-class Compactness $\downarrow$} \\
\cmidrule(lr){2-9}
& Head & Med & Tail & All 
& Head & Med & Tail & All \\
\midrule
LA~\cite{menon2020long} 
& 6.0 & 6.0 & 5.7 & 5.9 
& 28.6 & 28.8 & 29.6 & 29.0 \\

DBM-LA~\cite{son2024difficulty}
& \textbf{7.0} & 6.3 & 6.0 & 6.2 
& 31.0 & 31.9 & 33.1 & 32.0 \\
\rowcolor[HTML]{D9E2F3}
CILA-LA (Ours) 
& 6.8 & \textbf{6.9} & \textbf{6.1} & \textbf{6.6} 
& \textbf{28.1} & \textbf{28.3} & \textbf{29.3} & \textbf{28.6} \\
\bottomrule
\end{tabular}
\label{tab:dbm_analysis}
\vspace{-1em}
\end{table}

\noindent\textbf{Prediction Uncertainty.}
To verify that CILA mitigates NTLOS in terms of prediction confidence, we measure mean prediction entropy on test data across head, medium, and tail classes. 
As shown in Figure 5, LA consistently exhibits higher entropy across all class groups and imbalance factors, with particularly large values for tail classes, directly reflecting the suppressed non-target gradients identified in Section~\ref{sec:problem_NSD}. In contrast, CILA substantially reduces entropy across all settings, confirming that the confusion-aware adjustment restores informative gradients and yields more stable, confident predictions.

\begin{figure}[h] 
\begin{center}
\includegraphics[width=1\linewidth]{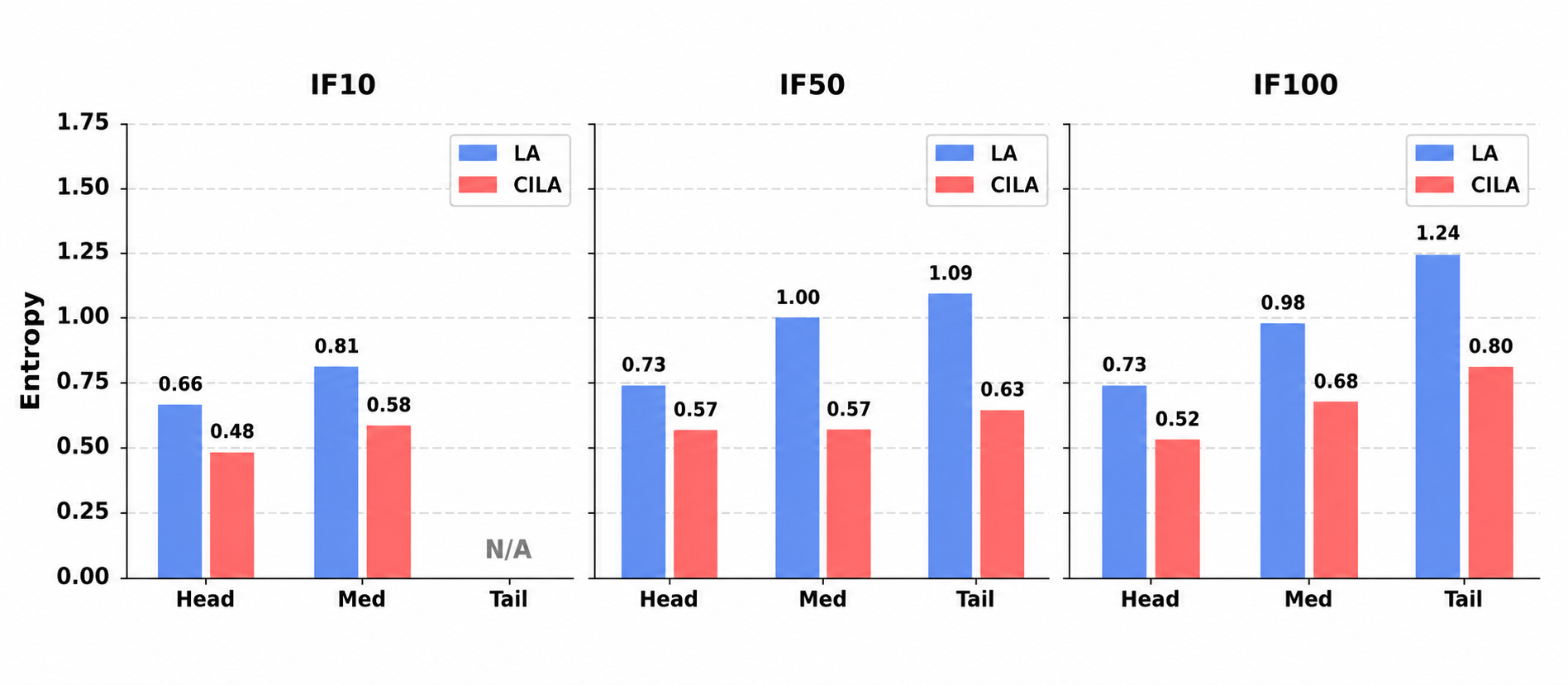}
\end{center}
\vspace{-2em}
\caption{Mean prediction entropy on test data.}
\vspace{-2em}
\end{figure}

\noindent \textbf{Effect of Batch-Adaptive Noise Scale.}
To evaluate the effect of adaptive noise scaling, we track the confusion penalty and accuracy during training under three settings: no noise~(zero), fixed-scale noise~(fix), and batch-adaptive noise~(mu), as shown in Figure 4.

Without noise, the tail penalty nearly vanishes after 100 epochs despite persistently low tail accuracy, confirming that many tail instances are correctly classified yet remain near the decision boundary. 
With batch-adaptive noise, these instances continue to receive meaningful penalties throughout training, leading to sustained improvement in tail accuracy. Compared to fixed-scale noise, batch-adaptive scaling maintains larger penalties on tail classes, as the noise scale is governed by the batch mean target logit, which is inherently smaller for tail classes. 
Consequently, batch-adaptive noise improves tail performance with minimal impact on head accuracy, demonstrating that the adaptivity of the scaling, not merely the presence of noise, is critical for tail generalization.

\noindent\textbf{Confusion between Semantically Similar Classes.}
To directly evaluate whether CILA improves discrimination between confusing classes, we analyze the confusion matrix on semantically similar class pairs. As shown in Figure 6, LA frequently misclassifies samples among visually similar categories, whereas CILA reduces such confusion, confirming that the instance-wise penalty effectively restores the learning signal needed to separate fine-grained classes.

\begin{figure}[h] 
\vspace{-2em}
\begin{center}
\includegraphics[width=1\linewidth]{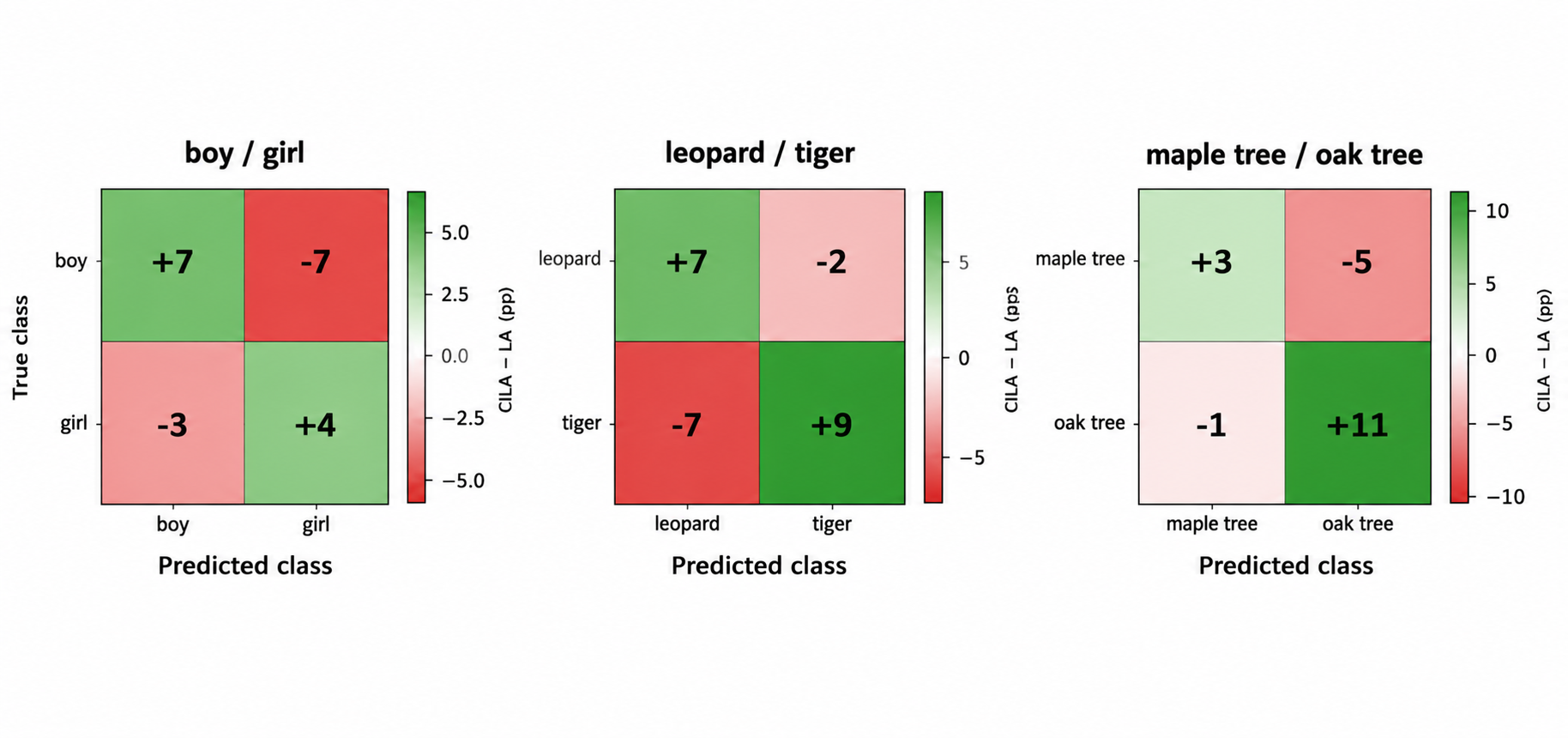}
\end{center}
\vspace{-3em}
\caption{Difference in confusion matrices (CILA $-$ LA) for
semantically similar class pairs on CIFAR-100-LT (IF=100).
Positive values~(green) indicate improved correct predictions
by CILA, confirming reduced misclassification between
fine-grained classes.}
\vspace{-2em}
\end{figure}

\subsection{ Ablation Study}
\label{sec:ablation}
We conduct ablation studies to validate the contribution of each component of CILA and the design choices of BNI.
Additional results are provided in the supplementary.

\noindent \textbf{Component Ablation.} 
Table~\ref{tab:aba_comp} summarizes the performance improvements obtained by progressively enabling CILA, BNI, and the class weighting term ($D_y$ in Eq.~\ref{eqn:class_wise}). Both CILA and BNI yield improvements over the baseline LA accuracy. However, when $D_y$ is omitted, accuracy improves for head classes but degrades for medium and tail classes, demonstrating that $D_y$ effectively balances learning across class groups and mitigates head-class dominance.

\begin{table}[h]
\caption{Ablation study on the three components under different imbalance factors (IF = 50, 100). \checkmark indicates the corresponding component is enabled.}
\renewcommand{\arraystretch}{1.2}
\setlength{\tabcolsep}{3.5pt}
\centering
\scriptsize
\begin{tabular}{ccc| cccc|cccc}
\toprule
\multicolumn{3}{c|}{Components} 
& \multicolumn{4}{c|}{IF 50} 
& \multicolumn{4}{c}{IF 100} \\
\midrule
CILA & BNI & $D_y$
& Head & Med & Tail & All
& Head & Med & Tail & All \\
\midrule
 &  &  
& 62.7 & 47.5 & 36.8 & 51.8
& 62.3 & 47.1 & 25.3 & 45.9 \\ \hline
\checkmark &  &  
& 64.6 & 46.8 & 36.1 & 52.1
& 64.9 & 47.0 & 25.8 & 46.9 \\
\checkmark &  & \checkmark  
& 64.3 & 47.0 & 37.3 & 52.3
& 64.6 &\textbf{ 47.6} & 28.0 & 47.6 \\ \hline
 & \checkmark &  
& 69.8 & 42.8 & 22.1 & 50.1
& 70.3 & 45.4 & 15.2 & 45.1 \\
 & \checkmark & \checkmark  
& 63.3 & 47.9 & 37.7 & 52.4
& 62.2 & 47.1 & 29.1 & 47.0 \\ \hline
\checkmark & \checkmark & 
& \textbf{70.0} & 42.4 & 21.3 & 49.9
&\textbf{ 71.4} & 44.8 & 13.5 & 44.7 \\
\rowcolor[HTML]{D9E2F3}
\checkmark & \checkmark & \checkmark
& 61.3 & \textbf{49.3} & \textbf{42.2} & \textbf{53.0}
& 64.7 & \textbf{47.6} & \textbf{30.1} & \textbf{48.3} \\
\bottomrule
\end{tabular}
\vspace{-1em}
\label{tab:aba_comp}
\end{table}

\noindent \textbf{Effect of Penalty Upper Bound.}
We analyze the sensitivity to the clipping upper bound of the confusion-based penalty. 
As shown in Table~\ref{tab:p_cliping}, smaller bounds such as 0.5 or 1 degrade performance by truncating informative penalty signals from strongly confusing instances, whereas larger bounds of 1.5 and 2 retain stronger correction signals. In particular, an upper bound of 2 yields the largest gains on tail-class accuracy, indicating that confusion signals for tail classes are preserved rather than truncated. 
We therefore adopt 2 as a robust operating point, which preserves informative confusion signals while preventing excessively large corrections from dominating the optimization.

\begin{table}[h]
\centering
\scriptsize
\caption{Performance comparison across penalty upper bound.}
\renewcommand{\arraystretch}{1.2}
\setlength{\tabcolsep}{4.5pt}
\vspace{-1em}
\begin{tabular}{c| cccc|cccc}
\toprule
\multirow{2}{*}{Upper Bound}
& \multicolumn{4}{c|}{IF 50} 
& \multicolumn{4}{c}{IF 100} \\
\cline{2-9}
& Head & Med & Tail & All
& Head & Med & Tail & All \\
\hline
0.5 
& 62.3 & 48.0 & 37.2 & 51.9
& 62.7 & 47.4 & 29.3 & 47.4\\
1.0 
& \textbf{63.2} & 47.0& 39.1& 52.2
& 64.3&	\textbf{48.4}& 28.1& 47.9\\
1.5 
& 64.3 & 47.4 & 37.3 &  52.5
& \textbf{64.8} & 48.2 & 28.8 & 48.2 \\
\rowcolor[HTML]{D9E2F3}
2.0
& 61.3 & \textbf{49.3} & \textbf{42.2} & \textbf{53.0}
& 64.7 & 47.6 & \textbf{30.1} & \textbf{48.3} \\
\bottomrule
\end{tabular} 
\label{tab:p_cliping}
\end{table}

\noindent\textbf{Effect of Noise Type.}

We further examine the choice of noise distribution in BNI. As shown in Table~\ref{tab:n_type1}, Rayleigh noise consistently achieves the best overall performance, with notable gains on tail classes.
This aligns with our design criterion that the injected noise should provide a non-trivial positive perturbation rather than concentrate near zero (Supp. Fig.1). 
Gaussian and Chi noise place substantial probability mass near zero, whereas Rayleigh has its mode at a positive value, more effectively exposing near-boundary samples to the confusion-aware penalty. These results support Rayleigh noise as a suitable choice for BNI.


\begin{table}[h]
\centering
\scriptsize
\setlength{\tabcolsep}{4.5pt}
\renewcommand{\arraystretch}{1.2}
\caption{Performance comparison across noise types.  }
\vspace{-1em}
\begin{tabular}{l| cccc|cccc}
\toprule
\multirow{2}{*}{Noise Type }
& \multicolumn{4}{c|}{IF 50} 
& \multicolumn{4}{c}{IF 100} \\

\cline{2-9}
& Head & Med & Tail & All
& Head & Med & Tail & All \\
\hline
Chi
& 63.3 & 48.4 & 35.5 & 52.2	
& 63.4 & 48.5 & 28.5 & 47.7\\
Gaussian 
& \textbf{65.0} & 47.0 & 39.0 & 52.9	
& 64.3 & \textbf{48.7} & 28.9 & 48.2 \\
\rowcolor[HTML]{D9E2F3}
Rayleigh  
& 61.3 & \textbf{49.3} & \textbf{42.2} & \textbf{53.0}
& \textbf{64.7} & 47.6 & \textbf{30.1} & \textbf{48.3}  \\
\bottomrule
\end{tabular} 
\label{tab:n_type1}
\vspace{-1em}
\end{table}

\noindent \textbf{Effect of noise ratio $\rho$.}
We further analyze the sensitivity of the noise ratio $\rho$, which controls the proportion of the mean target logit magnitude used to determine the injected noise scale (Eq.~\ref{eq:sigma}).
Based on our theoretical derivation, $\rho = 0.2$ is prescribed for CIFAR-100-LT. As confirmed in Table~\ref{tab:n_ratio}, this value consistently achieves the best overall accuracy across all imbalance factors, with particularly notable gains in tail classes, validating the theoretical grounding of our noise design.

\vspace{-1em}
\begin{table}[h]
\centering
\scriptsize
\setlength{\tabcolsep}{4pt}
\caption{Effect of noise ratio $\rho$ on accuracy.}
\vspace{-1em}
\begin{tabular}{c|  cccc|cccc}
\toprule
\multirow{2}{*}{$\rho$ in Eq.~\ref{eq:sigma} }
& \multicolumn{4}{c|}{IF 50} 
& \multicolumn{4}{c}{IF 100} \\
\cline{2-9}
& Head & Med & Tail & All
& Head & Med & Tail & All \\
\hline
0.1  
& 64.0 & 48.3 & 38.0 & 52.9
& \textbf{64.7} & 48.0 & 27.9 & 47.8  \\
\rowcolor[HTML]{D9E2F3}
0.2 
& 61.3 & \textbf{49.3} & \textbf{42.2} & \textbf{53.0}
& \textbf{64.7} & 47.6 & \textbf{30.1} & \textbf{48.3}  \\
0.3 
& \textbf{63.2} & 47.9 & 37.7 & 52.4
& 62.5 & \textbf{50.0} & 27.5 & 47.6\\
\bottomrule
\end{tabular} 
\label{tab:n_ratio}
\vspace{-2em}
\end{table}

\subsection{Computational Cost Analysis}  
On ImageNet-LT, using four RTX 4090 GPUs (batch:256), CILA (with BNI) takes 160 seconds per epoch compared to 153 seconds for LA, incurring only a 4.6\% overhead while achieving an 8.8\% absolute accuracy gain.
The added computation consists of lightweight element-wise operations on output logits, requiring no extra forward passes, gradients, or parameters. Thus, CILA achieves substantial performance gains at nearly the same training cost as LA.

\section{Conclusion}
We identified the Non-Target Logit Over-Suppression (NTLOS) problem in class-wise logit adjustment and formally characterized it at the gradient level. 
To address this, we proposed CILA, which restores informative gradients via a confusion-based penalty, and BNI, which extends the update to near-boundary samples via batch-adaptive noise injection, consistently improving performance across long-tailed benchmarks.


{
    \small
    \bibliographystyle{ieeenat_fullname}
    \bibliography{main}
}

\end{document}